\documentclass[runningheads]{llncs}
\usepackage{microtype}
\usepackage{graphicx}
\usepackage{subfigure}
\usepackage{array}
\usepackage{float}
\usepackage{multirow}
\usepackage{adjustbox}
\usepackage{booktabs}
\usepackage{hyperref}
\usepackage{booktabs} 
\usepackage{mathtools}
\usepackage{xcolor}

\usepackage{xr}
\makeatletter
\newcommand*{\addFileDependency}[1]{
  \typeout{(#1)}
  \@addtofilelist{#1}
  \IfFileExists{#1}{}{\typeout{No file #1.}}
}
\makeatother

\newcommand*{\myexternaldocument}[1]{%
    \externaldocument{#1}%
    \addFileDependency{#1.tex}%
    \addFileDependency{#1.aux}%
}
\myexternaldocument{supplementary}

\usepackage[T1]{fontenc}
\begin{document}
\title{Data Pruning via Separability, Integrity, and Model Uncertainty-Aware Importance Sampling}
%
%
\author{Steven Grosz\inst{1}\thanks{This author's contribution was performed as an intern with Amazon.} \and
Rui Zhao\inst{2} \and
Rajeev Ranjan\inst{2} \and
Hongcheng Wang\inst{2} \and
Manoj Aggarwal\inst{2} \and
Gerard Medioni\inst{2} \and
Anil Jain\inst{1}
}
\authorrunning{S. Grosz et al.}
%
\institute{Michigan State University, East Lansing MI 48824, USA \email{\{groszste,jain\}@msu.edu} \and
Amazon, Seattle, WA 98101, USA \email{\{zhaori,rvranjan,hongchw,manojagg,medioni\}@amazon.com}}
\maketitle              
\begin{abstract}
This paper improves upon existing data pruning methods for image classification by introducing a novel pruning metric and pruning procedure based on importance sampling. The proposed pruning metric explicitly accounts for data separability, data integrity, and model uncertainty, while the sampling procedure is adaptive to the pruning ratio and considers both intra-class and inter-class separation to further enhance the effectiveness of pruning. Furthermore, the sampling method can readily be applied to other pruning metrics to improve their performance. Overall, the proposed approach scales well to high pruning ratio and generalizes better across different classification models, as demonstrated by experiments on four benchmark datasets, including the fine-grained classification scenario.

\end{abstract}
\section{Introduction}
\label{Introduction}
The escalating size of models and datasets has led to an increase in the cost of training deep models. To address the challenges in maintaining high accuracy and scalability in the training process, various data reduction strategies have been proposed. These strategies aim to curate a smaller set of data that maximally retains information of the original dataset, facilitating more efficient learning of feature representations. Two overarching categories encapsulate most data reduction techniques: data pruning and data distillation. Data pruning involves identifying a subset of existing data for retention, whereas data distillation focuses on synthesizing a small number of new samples that lead to similar model accuracy as the original dataset~\cite{wang2018dataset}. 

While many data pruning methods are effective, there are remaining challenges that hinder the practical application of these techniques. One limitation of the existing pruning methods is the lack of robustness to data noise. For example, one proposed method utilizes the $l_2$ error of the predictions~\cite{paul2021deep} as a pruning metric, which is unable to accurately discriminate between difficult but useful samples and noisy, unusable samples~\cite{kim2022adaface}. This is due to the absence of a measure of sample quality, in addition to sample difficulty. 
%
Another drawback lies in the potential exacerbation of class imbalances by existing pruning strategies. The strict sampling approach, based on an increasing order of sample difficulty, may inadvertently lead to or intensify existing disparities between classes~\cite{pote2023classification}. While some prior methods suggest pruning an uneven number of samples per majority and minority classes to balance the number of samples~\cite{susan2021balancing}, such approaches overlook the inherent variations in difficulty among different classes. 
%
Moreover, the adaptability of existing methods is questionable, particularly concerning the decision of whether to prune difficult or easy samples. This decision hinges on factors such as the initial data volume and the chosen pruning ratio. The prevailing observation, as highlighted in \cite{sorscher2022beyond}, underscores the optimal strategy of retaining hard samples when the dataset is large and easy samples when it is small. To the best of our knowledge, few prior methods have embraced this insight in their pruning procedures. 

In this paper, we aim to design a scalable data pruning metric which improves upon the state-of-the-art by overcoming the previously mentioned limitations. Concretely, our contributions towards this goal are the following:

\begin{enumerate}
    \item A novel data pruning metric that intuitively captures multiple factors of data utility, including \textbf{S}eparability, \textbf{I}ntegrity, and \textbf{M}odel uncertainty (SIM). 
    \item An importance sampling procedure to prune data, which can be used with commonly used pruning metrics to improve the effectiveness of pruned data. 
    \item Evaluation on four benchmark datasets ranging from small to large number of classes show our method, which combines SIM with \textbf{S}ampling \textit{i.e.}, SIMS, has better scalability when the pruning ratios is high, better cross-model generalization, and reduced time needed to calculate pruning metrics than other approaches. 
\end{enumerate}

\section{Related Work}
\label{Related Work}
\subsection{Data Pruning}
One of the first pruning approaches proposed was Forgetting Scores~\cite{toneva2019empirical}, where samples were pruned based on the number of times they transitioned from being correctly to incorrectly classified throughout the training process of an expert model trained on the full dataset. Although performing well across many different classification datasets, Forgetting Scores suffer from high computational costs due to training the expert models to convergence on the full training dataset. To reduce this time spent computing pruning scores, Paul \textit{et al.} \cite{paul2021deep} proposed a new metric, EL2N, based on the magnitude of the error norm averaged across multiple expert models trained on the full dataset. Importantly, the authors showed that good performance could be achieved through training the expert models for only 20 epochs on the original datasets. However, a limitation present in both EL2N and Forgetting Scores is that they do not explicitly consider the recognizability of those samples in relation to their class centers in the embedding space. Similarly, other approaches have proposed using other techniques \textit{e.g.}, gradient of the loss~\cite{killamsetty2022automata,killamsetty2021grad}, generalization influence~\cite{yang2022dataset}, label noise~\cite{park2023robust}, moving-one-sample-out \cite{tan2023data}, \textit{etc.} But these methods also do not consider the inherent separability of different classes.

To address this limitation, Sorscher \textit{et al.} \cite{sorscher2022beyond} proposed Prototype and Self-Prototype scores for pruning. Here, the distance of each data sample from its corresponding class center is used as a pruning score, where samples farther from the class center are considered more challenging and are likely to be retained. With these simple metrics, the authors obtained similar pruning performance as the baseline EL2N and Forgetting Scores. However, two important limitations remain. First, neither of the previously mentioned pruning metrics explicitly models the quality of each data sample. This is important as without a measure of sample quality, the distinction between difficult, recognizable samples, and difficult, unrecognizable samples is missed. Secondly, even though Prototype scores consider the recognizability of each data sample in relation to it's own class center, they do not utilize the separability of samples compared to other class samples. Thus, information regarding the inherent separability of classes is overlooked. Our proposed metric aims to address each of the aforementioned challenges by explicitly measuring indications of data integrity, model uncertainty, and class separability in computing scores for pruning.

Another importance consideration in data pruning is not only the metric used to obtain pruning scores, but in how those scores are used to sample the original data for pruning. Many approaches sort the scores from low to high and retain the percentage of samples based on the highest scores. However, as noted in \cite{sorscher2022beyond}, the optimal sampling strategy depends on the amount of data retained. With a similar motivation, Xia \textit{et al.} \cite{xia2022moderate} proposed to sample scores in proximity to the median of each data class's distribution and showed improved results compared to using a strict sorting, especially at higher pruning ratios. Zheng \textit{et al.}
\cite{zheng2022coverage} also demonstrated the benefit of maintaining the coverage of data across different classes, which mitigates performance degradation at high pruning ratio.
In our work, we build upon these observations and propose an importance sampling procedure which depends on the given pruning ratio and allows us to vary the importance of in addition to the difficulty of the samples retained.

Finally, there is a line of work aimed at online pruning, where a certain percentage of the data is pruned away continually at some specified interval (\textit{e.g.}, every epoch, time elapsed, \textit{etc.})~\cite{chitta2021training,khan2023clip,killamsetty2023milo,li2018data}. 
These approaches are not directly comparable to ours since the pruning is applied continuously throughout the training process, rather than once at the start of training.

\subsection{Data Distillation}
Related to data pruning is the concept of data distillation. Rather than selecting a subset of images to retain, distillation aims to synthesize a set of representative samples to replace the original dataset. The idea was originally introduced in \cite{wang2018dataset} and showcased promising results in distilling small image datasets (\textit{e.g.}, MNIST, CIFAR-10) into just a single image per class. Subsequent works expanded on this by introducing label distillation, demonstrating the possibility of distilling datasets to less than one image per class \cite{sucholutsky2021less}. Ongoing research has focused on refining both data and label distillation through enhanced optimization functions for learning the distilled data. A major limitation of data distillation methods is that they suffer from computational issues and do not scale well to synthesizing a larger number of images per class in order to be useful in practice. Sundar \textit{et al.} \cite{sundar2023prune} proposed to alleviate the computational burden by first pruning the original dataset and then performing data distillation on the remaining samples. For an in-depth review of data distillation, refer to the recent survey~\cite{sachdeva2023data} and more recent methods such as \cite{hemultisize}. 


\section{Methods}
\label{Methods}

\begin{figure}
\begin{center}
\includegraphics[width=0.5\linewidth]{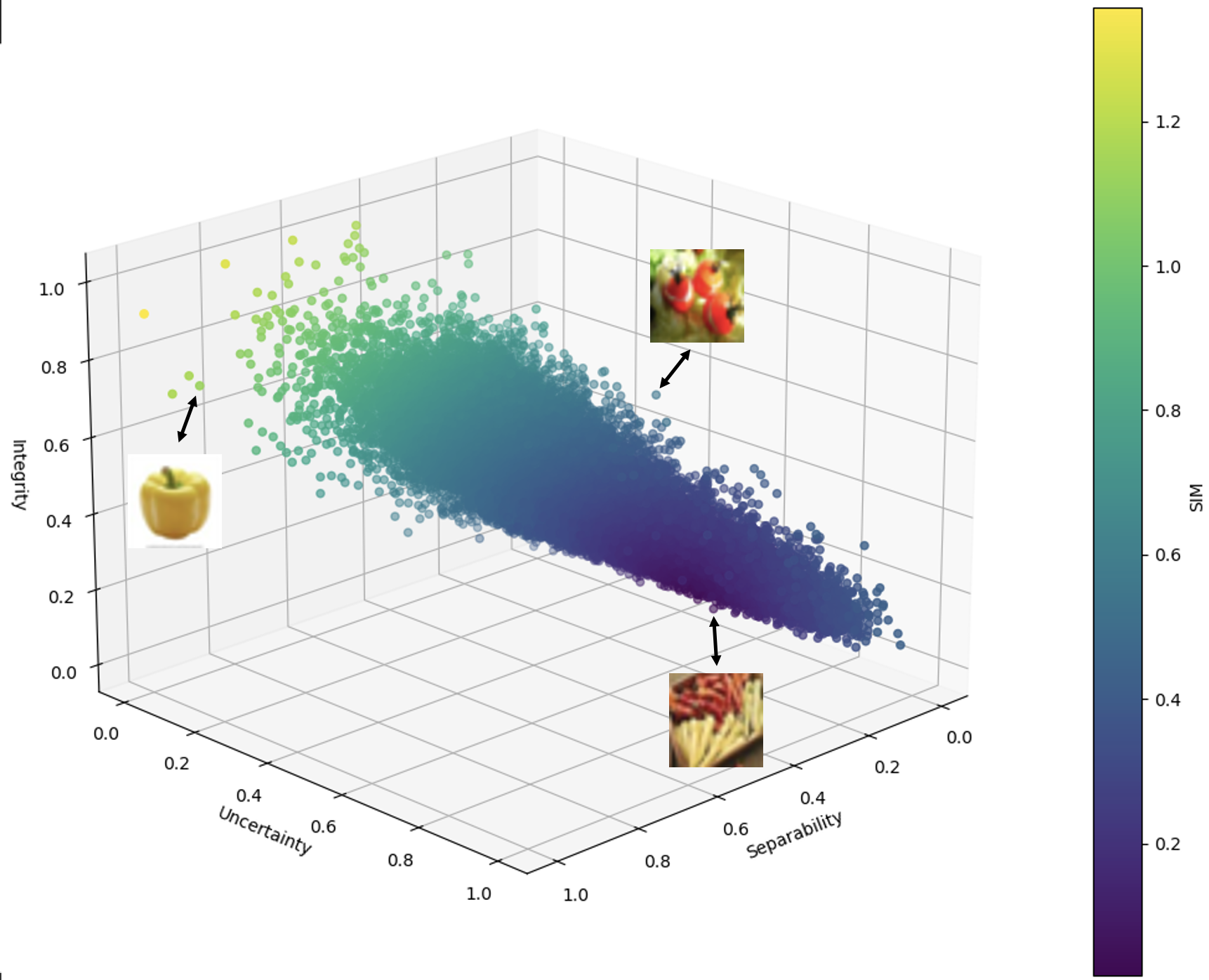}
\caption{Scatter plot of SIM scores for the CIFAR-100 dataset. The example images shown are from the ``sweet pepper" class and are provided to give a visualization of the typical samples falling in those respective regions of the graph.}
\label{fig:SIMS}
\vspace{-5em}
\end{center}
\end{figure}

\subsection{Problem Statement}
Consider learning a classification model $f_\theta(\cdot)$ parameterized by $\theta$ using dataset $\mathcal{D}=\{(x_i, y_i) \}_{i=1,...,|\mathcal{D}|}$, which consists of a collection of samples $x_i$ and corresponding class label $y_i\in \{1,...,C\}$ with $C$ being the number of classes. 
Denote $|\mathcal{D}|$ as the number of samples in the original dataset.
We aim to develop a method that can identify a subset $\mathcal{D}^\prime\subseteq\mathcal{D}$, such that $f_{\theta^\prime}(\cdot)$ learned from $\mathcal{D}^\prime$ yields comparable accuracy with $f_{\theta}(\cdot)$. The reduction of data can be quantified through a pruning ratio defined as $\alpha = 1-\mathcal{D}^\prime/\mathcal{D}$. Given $0 \le|\mathcal{D}^\prime| \le |\mathcal{D}|$, we have $\alpha\in [0,1]$.

\subsection{Data Separability}
Data separability or discriminability describes the inherent and unchangeable properties of a given data sample due to noise and strength of the information portrayed~\cite{duda2006pattern}. 
Many previous pruning metrics measure the difficulty of each data sample as the pruning score. These scores are then used to rank the samples in order to determine which ones should be removed from the dataset. In the case of EL2N, the sample difficulty is measured as the $l_2$ norm of the error: $\lVert\hat{y}_i - y_i\rVert$. For Prototype scores, the difficulty is measured as the distance of each data sample to its mean class embedding: $\lVert f_\theta(x_i) - \frac{1}{n}\sum^n_{j=1}{f_\theta(x_j)}\rVert$, where $n$ is the number of samples of corresponding class. However, neither of the measure considers the inter-class distance as an additional measure of sample difficulty. In case where all of the classes are well separated in the embedding space, this would not be an issue. However, when two or more classes are clustered together in the embedding space, then samples belonging to overlapping classes should be considered as more difficult than samples belonging to a perfectly separated class. For this reason, we choose to measure data separability using a modified version of the Recognizability Index (RI) \cite{chai2023recognizability}. Specifically, we define $d_\theta^P(x_i) = 1 - \cos(\phi_{y_i})$ as the positive distance of a sample $x_i$ to its actual class center under model $\theta$, where $\phi_{y_i}$ is the absolute angle between $f_\theta(x_i)$ and center embedding of class $y_i$. 
We then define $d_\theta^N(x_i) = 1 - \max_{j \in \{1, \ldots, C\} \backslash \{y_i\}} \cos(\phi_{j})$ as the negative distance of a sample to its nearest imposter class under model $\theta$, where $\phi_j$ is the absolute angle between $f_\theta(x_i)$ and center embedding of class $j$. 
For both $d_\theta^P$ and $d_\theta^N$, the center embedding is calculated as the average embeddings of class samples. 
Finally, data separability is computed as follows:
\begin{align}
s_\theta(x_i) &= \frac{d_\theta^N(x_i)}{d_\theta^P(x_i) + \varepsilon}
\end{align}
where $\epsilon=1e^{-7}$ is a small constant to avoid division by 0. Intuitively speaking, $s(x_i)$ reflects the difficulty to classify the data sample. An easily recognizable sample has small positive distance and large negative distance, thus a high value of $s(x_i)$. Otherwise $s(x_i)$ is low.

\subsection{Data Integrity}
A significant shortcoming of existing data pruning metrics is the failure to consider the integrity (\textit{i.e.}, quality) of data samples. For a challenging sample, without capturing sample quality, the important distinction between recognizable and unrecognizable sample is missed. This oversight may lead to discarding hard but useful samples (\textit{e.g.}, profile-view of an object), instead of difficult samples due to low quality (\textit{e.g.}, blurred image).
To capture quality, we use the embedding norm as it has been previously shown to be correlated with sample quality~\cite{kim2022adaface} and is inexpensive to compute. In particular, given a classification model $f_\theta(\cdot)$, the embedding norm is defined as the $l_2$-norm of the embedding vector. 
\begin{align}
e_\theta(x_i) = || f_\theta(x_i) ||_2
\end{align}
Empirically, it is observed that higher quality $x_i$ often yields higher value of $e_\theta(x_i)$ and vice versa for a given model $\theta$.

\subsection{Model Uncertainty}
So far, we have been focusing on deriving pruning metrics based on information possessed by the data itself. In practice, whether a model can utilize such information also impacts the effectiveness of pruned data. Therefore, we incorporate model uncertainty in our pruning metric, which is computed by Jensen-Shannon Divergence (JSD) of a set of expert models predictions $f(x_i)$ on a given data sample $x_i$. Similar approach was also adopted in prior work such as \cite{lakshminarayanan2017simple,shoshan2024asymmetric}.
As defined in Eq. (\ref{eq:JSD}), low uncertainty estimates are achieved when the entropy between individual model probabilistic distributions is low, meaning that all the models assign a similar probability distribution to different classes for given $x_i$. On the other hand, high entropy between probabilistic distributions on $x_i$ indicates less agreement between the expert models on how to classify a given image example.
\begin{align}
\label{eq:JSD}
JSD(\{h_j\}_{j=1}^K\mid x_i) &= H(M) - \frac{1}{K}\sum_{j=1}^{K} H(h_j(x_i)) 
\end{align}
where $M = \frac{1}{K}\sum_{j=1}^{K} h_j(x_i)$ is the average probability distribution of $x_i$ and $K$ is the number of expert models. $H(h)$ is the Shannon entropy for distribution $h$. In practice, the expert models can be obtained by training different models using the entire dataset. $h$ can be estimated as the softmax output of classification model.
Intuitively speaking, more noisy and lower quality samples lead to higher uncertainty. 
Notice that $JSD(\cdot)$ is bounded between 0 and 1. For consistency with separability and integrity, we denote the compliment of the uncertainty as $c(x_i)$, such that higher value of $c(x_i)$ represents higher certainty and vice versa. $c(x_i)$ is computed as follows, which is also bounded between 0 and 1.
\begin{align}
c(x_i) &= 1 - JSD(\{h_j\}_{j=1}^K\mid x_i)
\end{align}
\vspace{-3em}

\begin{figure*}
\begin{center}
\includegraphics[width=\linewidth]{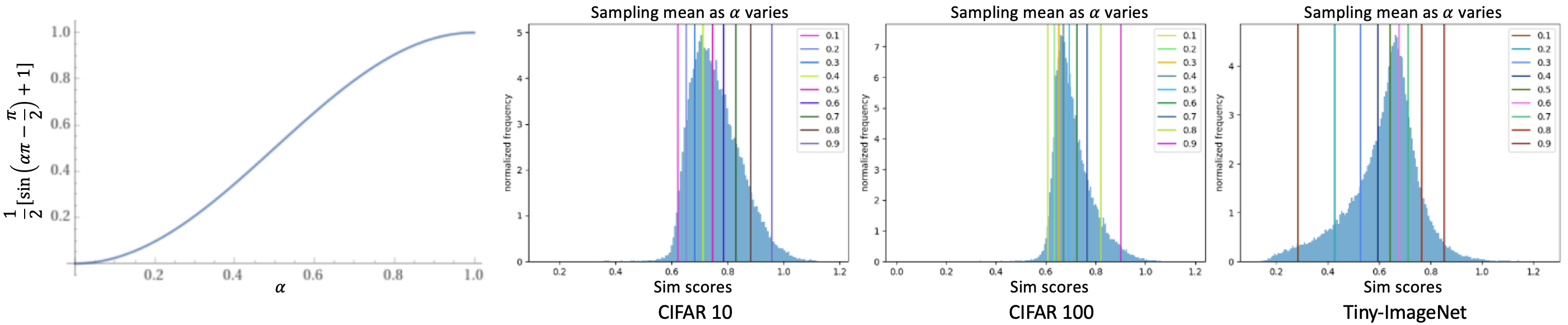}
\caption{Illustration of the sampling distribution mean at different pruning ratio $\alpha$.}
\vspace{-3em}
\label{fig:sampling_mean}
\end{center}

\end{figure*}

\begin{figure*}
\begin{center}
\includegraphics[width=\linewidth]{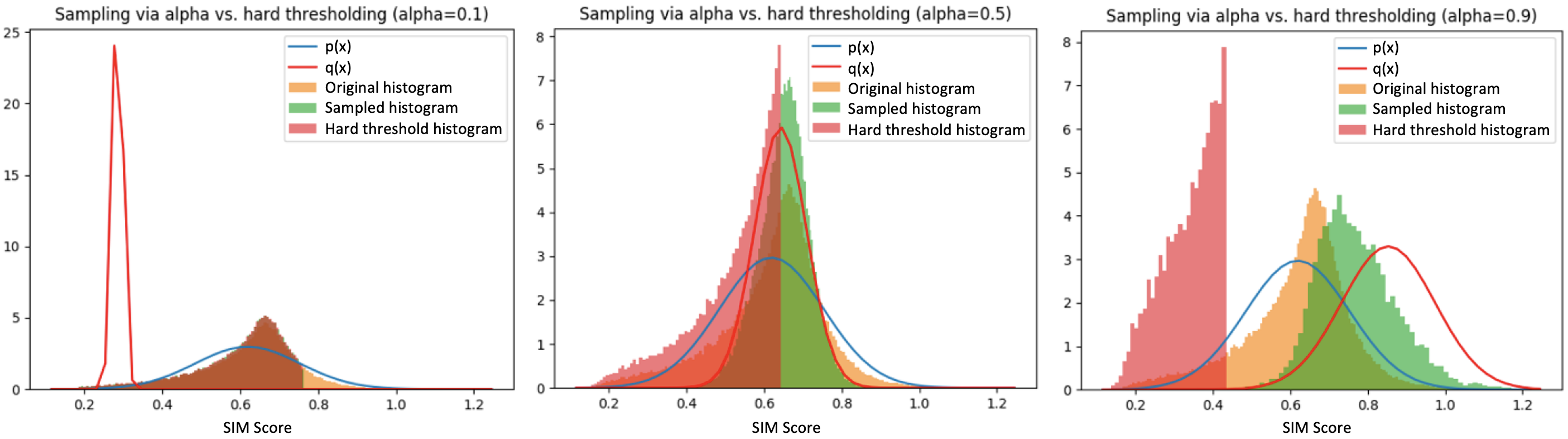}
\caption{Illustration of varying sampling distributions for Tiny-ImageNet dataset as pruning ratio $\alpha$ increases.}
\label{fig:varying_alpha}
\vspace{-2em}
\end{center}
\end{figure*}

\subsection{Derivation of SIM}
We now discuss how to combine these three metrics together as the proposed SIM metric. First, we leverage the same set of $K$ expert models used in uncertainty estimation to obtain aggregated separability and integrity defined as follows. 
\begin{align}
s(x_i) &= \frac{1}{K}\sum_{j=1}^{K} s_{\theta_j}(x_i) \\
e(x_i) &= \frac{1}{K}\sum_{j=1}^{K} e_{\theta_j}(x_i)
\end{align}
Then we normalize each metric $s(x_i), e(x_i)$, and $c(x_i)$ separately by subtracting the minimum followed by dividing by the range so that each metric is spread between 0 and 1. 
Finally, we combine the three metrics in the following way:
\begin{align}
    \label{eq:SIM}
    g(x_i) &=\sqrt{(1-s(x_i))^2+c(x_i)^2} - \sqrt{(1-s(x_i))^2+(1-c(x_i))^2} \\ 
    SIM(x_i)&=\sqrt{g(x_i)^2+e(x_i)^2}
\end{align}
The motivation for this formulation can be explained by the geometric interpretation of the two terms in Eq.~(\ref{eq:SIM}). 
The first term, $\sqrt{(1-s(x_i))^2+c(x_i)^2}$ can be seen as the distance to the top-right corner of the normalized separability-uncertainty plane. This term is high for low data separability and low uncertainty samples. Samples falling in this region are assumed to be hard to recognize but potentially useful given that the uncertainty is low. 
The second term, $\sqrt{(1-s(x_i))^2+(1-c(x_i))^2}$ can be viewed as the distance to the bottom-right corner of the normalized separability-uncertainty plane. This term is penalizing samples with low separability and high uncertainty. 
Thus, the first term is trying to give a positive weighting towards retaining those difficult, but recognizable samples and the second term is giving a negative weighting towards low separability and high uncertainty samples. The distribution of SIM scores is illustrated for the CIFAR-100 dataset in Fig.~\ref{fig:SIMS}. We also tried the formulation of $g'(x_i)=\sqrt{s(x_i)^2 + c(x_i)^2}$ but the performance is worse compared to $g(x_i)$.

\begin{figure}
\begin{center}
\includegraphics[width=\linewidth]{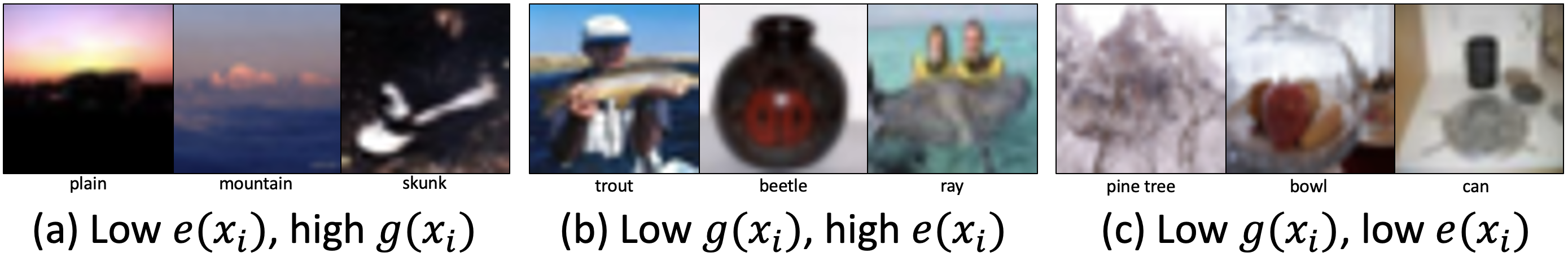}
\caption{Example images from CIFAR-100 with poor qualities such as blurriness, low contrast, and multiple classes. The advantage of computing SIM scores is evident in the different quality aspects captured by $e(x_i)$ and $g(x_i)$, where many difficult but useful samples would be rejected if either metric was used alone.}
\label{fig:low_SIM}
\end{center}
\vspace{-3em}
\end{figure}

An intuition as to why we combine each of these three metrics is shown in Fig.~\ref{fig:low_SIM}, which shows some example images from CIFAR-100 that have poor quality aspects that are captured by either $g(x_i)$, $e(x_i)$, or both. For example, according to the embedding norm $e(x_i)$, the images in subfigure (a) should have low scores due to the presence of blurry, low contrast images. Similarly, according to $g(x_i)$, the images in subfigure (b) should also have low scores due to the presence of multiple classes in the image. However, in all of these images, it seems reasonable to expect the classifier to learn to correctly identify the main objects. Indeed, if we had just used either $g(x_i)$ or $e(x_i)$ alone, then each of the images would have been likely pruned away early on from the training dataset. In contrast, by combining both $g(x_i)$ and $e(x_i)$, the resulting SIM scores actually fall towards the median and are retained for moderate values of $\alpha$. Finally, we also retain the ability to reject cases of unrecognizable samples that exhibit degradations captured by both $g(x_i)$ and $e(x_i)$, like those in subfigure (c) of Fig.~\ref{fig:low_SIM}, which are likely not contributing positively towards the learning task.

\subsection{SIM with Importance Sampling (SIMS)}
Existing pruning methods usually sort samples by the pruning metric and retain the samples that meet a pre-defined threshold. Despite its simplicity, retaining the most difficult samples according to the pruning metric may lead to under-fitting for large pruning ratios and datasets with more challenging classes. To address this limitation, we propose an adaptive sampling strategy based on the pruning ratio. It can be combined seamlessly with SIM and any other scalar-based pruning metric. We name our approach as SIMS, which consists of two key designs: i) varying the sample difficulty of pruned samples based on the pruning ratio and ii) sampling via a combination of class dependent 
and class independent distributions 
in order to avoid exacerbating bias in class-imbalanced data distributions.

\subsubsection{Importance Sampling based on Pruning Ratio}
\label{sec:is-base}
Our sampling strategy employs importance sampling~\cite{kloek1978bayesian} to emphasize the desired difficulty of samples during pruning. Rather than ranking the SIM scores from lowest to highest and retaining a percentage of the highest scoring samples based on the pruning ratio ($\alpha$), the data to retain is selected via a sampling procedure. We construct a sampling distribution $q(x|\alpha)$, that is dependent on $\alpha$, to assign an \textit{importance} weight $w(x|\alpha)$ to each data sample $x$ in the original data distribution $p(x)$ to bias the sampling based on their \textit{importance}. The \textit{importance} is defined as the ratio between $q(x|\alpha)$ and $p(x)$ (\textit{i.e.}, $w(x|\alpha)=\frac{q(x|\alpha)}{p(x)}$). Our intuition is that for small $\alpha$, a large percentage of difficult samples can be retained because we have a lot of data samples remaining, whereas for large $\alpha$, we have very few data samples retained so we need to retain the easy, representative examples.

In our implementation, both the original distribution $p(x)$ and sampling distribution $q(x|\alpha)$ assume the form of Normal distributions \textit{i.e.}, $p(x) \sim \mathcal{N}(\mu_0,\,\sigma_0^{2})$ and $q(x|\alpha) \sim \mathcal{N}(\mu,\,\sigma^{2})$, where $\mu_0$ and $\sigma_0$ are estimated from the training data directly. $\mu$ and $\sigma$ are estimated based on $\alpha$, $\mu_0$, and $\sigma_0$. Specifically,
$\mu$ is determined by Eq.~(\ref{eq:mu}), where $F^{-1}$ is the inverse cumulative distribution function of $p(x)$ and $t$ is the 
quantile function defined by Eq.~(\ref{eq:quant}). Notice $t$ is a sinusoidal function of $\alpha$, which increases monotonically with $\alpha$. As illustrated in Fig.~\ref{fig:sampling_mean}, smaller $\alpha$ reduces $\mu$ further so that $w(x|\alpha)$ is larger for $x$ with smaller pruning score \textit{i.e.} more difficult samples and vice versa.
\begin{align}
    \label{eq:quant}
    t &= \frac{1}{2}\left(\sin(\alpha \pi - \frac{\pi}{2}) + 1\right) \\
    \label{eq:mu}
    \mu &= F^{-1}\left(t; \mu_0, \sigma_0\right)
\end{align}
For $\sigma$, we simply use a scaled version of $\sigma_0$ as defined in Eq.~(\ref{eq:sigma}). This is based off the observation that most of the high SIM scores for each class (see Fig.~\ref{fig:ex_imgs} in supplementary) contain similar, redundant samples. Therefore, as $\alpha$ increases, $\sigma$ also increases so that we allow for extra diversity in the sampled images as the number of samples to retain becomes small.
\begin{align}
\label{eq:sigma}
    \sigma &= \alpha\sigma_0
\end{align}
An illustration of the importance sampling procedure is given in Fig.~\ref{fig:varying_alpha} for Tiny-ImageNet \cite{le2015tiny} using SIM scores and sampling at various pruning ratios. 

\subsubsection{Combination of Class-Dependent and Class-Independent Sampling}
\label{sec:is-improved}
Our decision to include both class-dependent and class-independent sampling is motivated from the following two observations. 
First, we observed that the performance of pruning using the Prototype score can be improved by reserving a percentage of the pruned dataset for samples at the center of each cluster. In particular, CIFAR-10 has the best performance when 10\% of the data is sampled from the class centers and 90\% from the boundary (see Table \ref{tab:proto} in the supplementary). Second, previous research has shown that data pruning via an overall score distribution for the entire dataset may exacerbate class imbalances~\cite{sorscher2022beyond}. Motivated by these observations, we adapt our importance sampling strategy to first sample a percentage of samples within each class (class-dependent sampling) and then sample the remaining percentage from the overall data distribution (class-independent sampling). Through an empirical analysis, the optimal performance was achieved by first sampling 5\% of samples from each individual class and the remaining 95\% from the overall data distribution.

\section{Experimental Results}
\label{Experimental Results}
In this section, we provide a comparison of the proposed SIMS approach with various benchmark methods. We report the comparison considering both factors of classification accuracy and training time on the pruned datasets. We also explore the generalization performance of SIMS when models used for deriving pruning metric and performing classification are different. More qualitative analysis of pruning results is provided in the supplementary.


\subsection{Datasets}
\label{sec:Datasets}
We conducted our experiments using four image classification datasets. CIFAR-10~\cite{krizhevsky2009learning}, CIFAR-100~\cite{krizhevsky2009learning}, and Tiny-ImageNet~\cite{le2015tiny} are well-known selections utilized in prior studies. CIFAR 10 and CIFAR 100 have 50,000 images with 10 and 100 classes, respectively, and Tiny-ImageNet has 100,000 images distributed equally among 200 classes. Finally, the large-scale, fine-grained dataset iNaturalist~\cite{van2018inaturalist} has 675,170 images with a long-tailed distribution across 5,089 classes. 

\begin{table*}
\renewcommand{\arraystretch}{1.3}
\centering
\caption{Classification accuracy on CIFAR-10, CIFAR-100, Tiny-ImageNet, and iNaturalist datasets. Results are averaged across 3 repeated experiments. Best result in each column for each dataset is in \textbf{bold}.}
\setlength{\tabcolsep}{3.5pt}
\begin{adjustbox}{width=\textwidth,center}
\begin{tabular}{|c|lccccccccc|}
\cline{1-11}
Dataset & Pruning Method & $\alpha=0.1$ & $\alpha=0.3$ & $\alpha=0.5$ & $\alpha=0.7$ & $\alpha=0.9$ & Avg. $\alpha=0.1$-$0.5$ & Avg. $\alpha=0.5$-$0.9$ & Avg. $\alpha=0.1$-$0.9$ & \\
\cline{1-11}
\multirow{7}{*}{\rotatebox{90}{CIFAR-10}} & Random & 93.14 & 92.23 & 90.52 & 87.45 & \textbf{72.17} & 92.05 & \textbf{84.88} & \textbf{88.24} & \\
 & EL2N & \textbf{93.50} & 93.35 & 92.57 & 83.12 & 27.88 & 93.23 & 69.25 & 79.98 & \\
 & Prototype & 93.31 & 93.19 & 92.43 & 87.68 & 48.28 & 93.10 & 79.25 & 85.48 & \\
 & Forgetting & 93.46 & \textbf{93.40} & \textbf{93.04} & 85.56 & 46.46 & \textbf{93.40} & 75.92 & 83.73 & \\
 \cline{2-11}
 & SIM & \textbf{93.50} & 93.33 & 92.46 & \textbf{88.42} & 42.26 & 93.22 & 77.84 & 84.76 & \\
 & SIMS & 93.42 & 93.20 & 91.18 & 85.34 & 69.35 & 92.83 & 83.01 & 87.56 & \\
\cline{2-11}
 & Full Dataset & \multicolumn{8}{c}{\textit{93.57 $\pm$ 0.18}} &  \\
\cline{1-11}
\multirow{7}{*}{\rotatebox{90}{CIFAR-100}} & Random & 71.39 & 68.85 & 64.48 & 55.96 & 28.49 & 68.26 & 51.28 & 59.25 & \\
 & EL2N & 72.25 & 69.04 & 51.79 & 23.42 & 6.37  & 65.50  & 26.31 & 45.25 & \\
 & Prototype & 72.45 & 68.96 & 58.40  & 35.96 & 7.52  & 67.09 & 33.97 & 49.66 & \\
 & Forgetting & 72.43 & \textbf{70.80}  & 63.62 & 50.1  & 21.72 & 69.53 & 46.27 & 57.27 & \\
 \cline{2-11}
 & SIM & \textbf{72.66} & 69.91 & 61.21 & 40.64 & 11.77 & 68.47 & 38.37 & 52.55 & \\
 & SIMS & 72.24 & 70.70  & \textbf{66.00}    & \textbf{58.11} & \textbf{33.83} & \textbf{69.83} & \textbf{53.88} & \textbf{61.39} & \\ 
\cline{2-11}
 & Full Dataset & \multicolumn{8}{c}{\textit{72.57 $\pm$ 0.06}} &  \\
\cline{1-11}
\multirow{7}{*}{\rotatebox{90}{Tiny-ImageNet}} & Random & 58.34 & 55.36 & 50.77 & 43.90  & 25.51 & 54.93 & 41.43 & 47.89 & \\
& EL2N & 59.30  & 54.14 & 40.58 & 21.15 & 7.65  & 52.04 & 22.78 & 37.06 & \\
& Prototype & 58.84 & 54.69 & 44.26 & 24.90  & 3.98  & 52.96 & 24.19 & 37.95 & \\
& Forgetting & 59.21 & \textbf{58.04} & \textbf{53.47} & 44.23 & 25.11 & \textbf{57.30}  & 42.18 & \textbf{49.33} & \\
\cline{2-11}
& SIM & 59.24 & 54.72 & 45.12 & 29.22 & 6.39  & 53.62 & 27.27 & 39.92 & \\
& SIMS & \textbf{59.79} & 55.49 & 51.72 & \textbf{45.53} & \textbf{27.60}  & 55.91 & \textbf{42.99} & 49.20 & \\
\cline{2-11}
& Full Dataset & \multicolumn{8}{c}{\textit{59.52 $\pm$ 0.41}} &  \\
\cline{1-11}
\multirow{5}{*}{\rotatebox{90}{iNaturalist}} & Random & 42.16 & 39.58 & 35.42 & 29.21 & 17.61 & 39.16 & 27.93 & \textbf{33.33} & \\
 & Forgetting & \textbf{43.61} & \textbf{40.99} & 32.33 & 17.78 & 4.57 & \textbf{39.61} & 18.19 & 28.52 & \\
 \cline{2-11}
 & SIM & 38.61 & 26.99 & 17.89 & 8.27 & 2.10 & 27.60 & 9.15 & 18.43 & \\
 & SIMS & 39.33 & 30.78 & \textbf{36.72} & \textbf{34.53} & \textbf{22.42} & 34.81 & \textbf{32.27} & 33.19 & \\
\cline{2-11}
 & Full Dataset & \multicolumn{8}{c}{\textit{42.90 $\pm$ 0.10}} &  \\
\cline{1-11}
\end{tabular}
\end{adjustbox}
\vspace{-2.5em}
\label{tab:class_acc}
\end{table*}

\subsection{Classification Accuracy}
\label{sec:acc}
We assessed various baseline pruning metrics: Random, EL2N \cite{paul2021deep}, Prototype \cite{sorscher2022beyond}, and Forgetting Scores \cite{toneva2019empirical} against the proposed SIMS
on datasets mentioned in Section \ref{sec:Datasets}. We vary $\alpha$ from 0.1 to 0.9 with 0.1 increments for each experiment. For fair comparison across different methods, we used ResNet18 models for both  calculating pruning metrics and classification for all methods. Furthermore, all experiments are conducted three times and the average results are reported. From the results in Table~\ref{tab:class_acc} and Table~\ref{tab:forget_sampling}, we make the following observations.

First, SIMS outperforms the next best method, Forgetting Scores, on CIFAR-10, CIFAR-100, and iNaturalist datasets and is comparable on Tiny-ImageNet. Next, the biggest difference in performance is observed in the high $\alpha$ regime \textit{i.e.}, $\alpha\in\{0.5,...,0.9\}$, where SIMS outperforms Forgetting Scores across all datasets by a large margin. For example, the models trained on SIMS pruned datasets achieved, on average, a 22.1\% increase in test classification accuracy over those trained on Forgetting Scores pruned datasets. Lastly, the incorporation of importance sampling substantially elevated the performance of both SIM (\textit{i.e.} SIMS without importance sampling) and Forgetting Scores, increasing the average performance across the four datasets for $\alpha\in \{0.1,...,0.9\}$ by 30.87\% and 8.57\% for SIM and Forgetting Scores, respectively. Thus, importance sampling is proved to be a useful strategy to pair with a general pruning metric in order to improve its performance.

Interesting to note is the high performance of random pruning averaged across all values of $\alpha$ for the iNaturalist dataset, although SIMS still performs significantly better for high pruning ratios ($\alpha\in \{0.5,...,0.9\}$). We posit that given the relatively low performance and difficulty of this fine-grained classification dataset, the performance at low pruning ratios actually degrades rapidly when biasing the selection to retain low SIM score samples. This is because the classification problem is already difficult enough, that focusing on the boundary samples for each class (low SIM scores), rather than the representative samples (high SIM scores), is hurting the performance significantly. Thus, it seems that biasing the original focus to easier or more challenging samples depending on the initial full dataset performance may prove as a useful exploration for future work, especially on more challenging long-tailed and fine-grained datasets.

\vspace{-1em}
\begin{table}
\centering
\caption{Accuracy of Forgetting vs. Forgetting plus importance sampling across CIFAR-10, CIFAR-100, Tiny-ImageNet, and iNaturalist datasets, averaged across $\alpha$ from 0.1-0.9.}
\begin{tabular}{lcccc}
\hline
Metric & CIFAR-10 & CIFAR-100 & Tiny & iNaturalist \\
\hline
Forgetting & 83.73 & 57.27 & \textbf{49.33} & 28.52 \\
Forgetting + Sampling & \textbf{87.82} & \textbf{61.23} & 47.12 & \textbf{36.04}\\
\hline
\end{tabular}
\vspace{-3em}
\label{tab:forget_sampling}
\end{table}

\begin{figure}
\begin{center}
\includegraphics[width=1.0\linewidth]{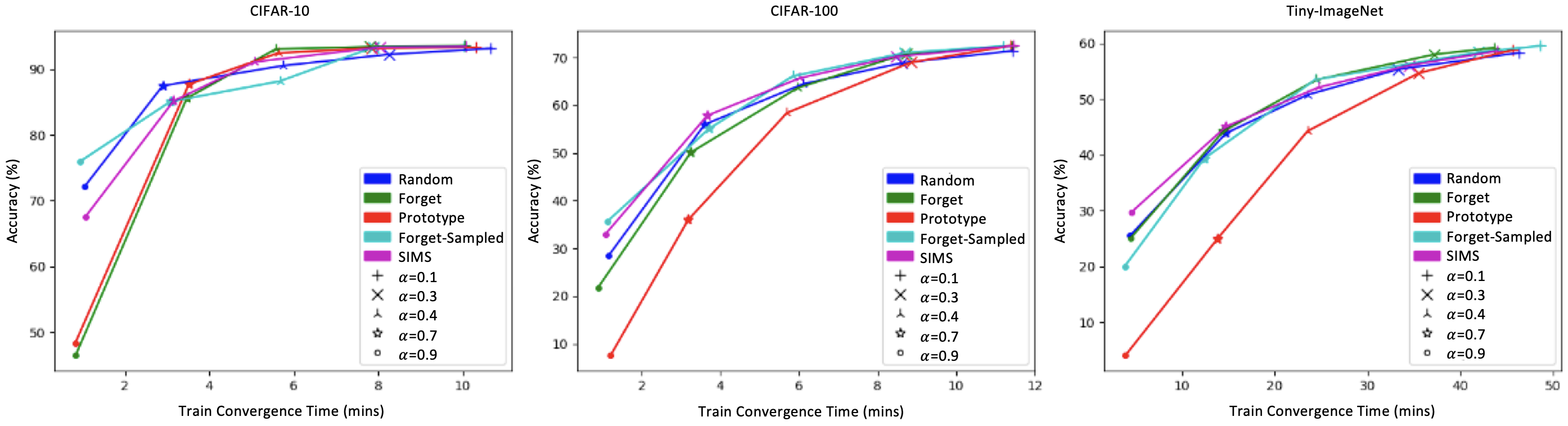}
\caption{Classification accuracy vs. training time of different pruning methods (Best viewed in color).}
\label{fig:time_vs_acc}
\end{center}
\vspace{-3em}
\end{figure}

\subsection{Training Time vs. Accuracy}
One benefit of data pruning is reducing the training time compared to the full dataset, which is beneficial for tasks like hyper-parameter tuning and neural architecture search. Fig.~\ref{fig:time_vs_acc} presents a plot of training time vs. testing accuracy for different dataset using different pruning metrics at different pruning ratios. Comparing SIMS to the next best benchmark method, Forgetting Scores, we see a clear separation in terms of accuracy at higher pruning ratios (left half of the curve) with comparable training time. Where the efficiency advantage of different pruning metrics becomes evident is in the time required to compute the metrics, where we only train each expert model for 20 epochs for computing SIM scores, compared to Forgetting Scores which requires at least 75 epochs for each expert model~\cite{toneva2019empirical}). Similarly, EL2N scores only require training for 20 epochs~\cite{paul2021deep}, but their accuracy is much lower across all of the datasets used in our evaluations.


\begin{table}
\centering
\caption{Accuracy of all possible combinations of metrics and sampling procedures. The best result in each column is in \textbf{bold}.}
\scriptsize
\begin{tabular}{lcccc}
\hline
Methods Ablation & CIFAR-10 & CIFAR-100 & Tiny & Avg. \\
\hline
S & 87.53 & 60.12 & 45.40 & 64.35 \\
I & 86.57 & 60.33 & 46.22 & 64.37 \\
M & \textbf{88.04} & 58.83 & 46.51 & 64.46 \\
S + I & 86.40 & 56.93 & 43.53 & 62.29 \\
S + M & 87.83 & 60.44 & 46.21 & 64.83 \\
I + M & 86.30 & 60.41 & 42.55 & 63.09 \\
S + I + M (proposed) & 87.56 & \textbf{61.39} & \textbf{49.20} & \textbf{66.05} \\
\hline
No sampling & 84.76 & 52.55 & 39.92 & 59.08\\
0\% class sampling & \textbf{87.82} & 61.23 & 47.12 & 65.39\\
1\% class sampling & 87.31 & 61.20 & 49.21 & 65.91\\
5\% class sampling (proposed) & 87.56 & \textbf{61.39} & 49.20 & \textbf{66.05}\\
100\% class sampling & 87.26 & 61.02 & \textbf{49.22} & 65.83\\
\hline
\end{tabular}
\label{tab:sims_ablation}
\end{table}


\begin{table}
\centering
\caption{Cross-model generalization. Each column represents different classification models and datasets. Each row represents different pruning metric and pruning model is ResNet18.}
\scriptsize
\setlength{\tabcolsep}{3.5pt}
\begin{tabular}{lcccc}
\hline
\multirow{2}{*}{Pruning metric} & \multicolumn{2}{c}{ResNet18} & \multicolumn{2}{c}{ResNet50} \\
& CIFAR-10 & CIFAR-100 & CIFAR-10 & CIFAR-100 \\
\hline
Forgetting Scores & 87.82 & 61.23 & 91.72 & 73.76 \\
SIMS & 87.31 & 61.20 & 92.09 & 74.17 \\
\hline
\end{tabular}
\label{tab:generalization}
\end{table}

\subsection{Ablation Analysis}
\label{sec:ablation}
An in-depth ablation study was conducted on the individual components of SIMS. For all studies in this section, we used $K=10$ expert ResNet18 models for pruning and used a different ResNet18 model trained on the pruned datasets for classification. 
Same as Section \ref{sec:acc}, we vary $\alpha$ from 0.1 to 0.9 with 0.1 increments and report average accuracy on different datasets. First, we ablated on the various combinations of the three metric components of SIM scores, including separability (S), integrity (I), and model uncertainty (M). The results presented in Table~\ref{tab:sims_ablation} upper subsection show that the best average performance of 66.05\% is obtained in the last row from the combination of all three metrics \textit{i.e.} SIM. We also notice the best performing individual metric is model uncertainty. And combining two metrics does not necessarily improve overall performance \textit{e.g.} S+I perform worse than S or I used separately.

\begin{figure}
\begin{center}
\includegraphics[width=0.75\linewidth]{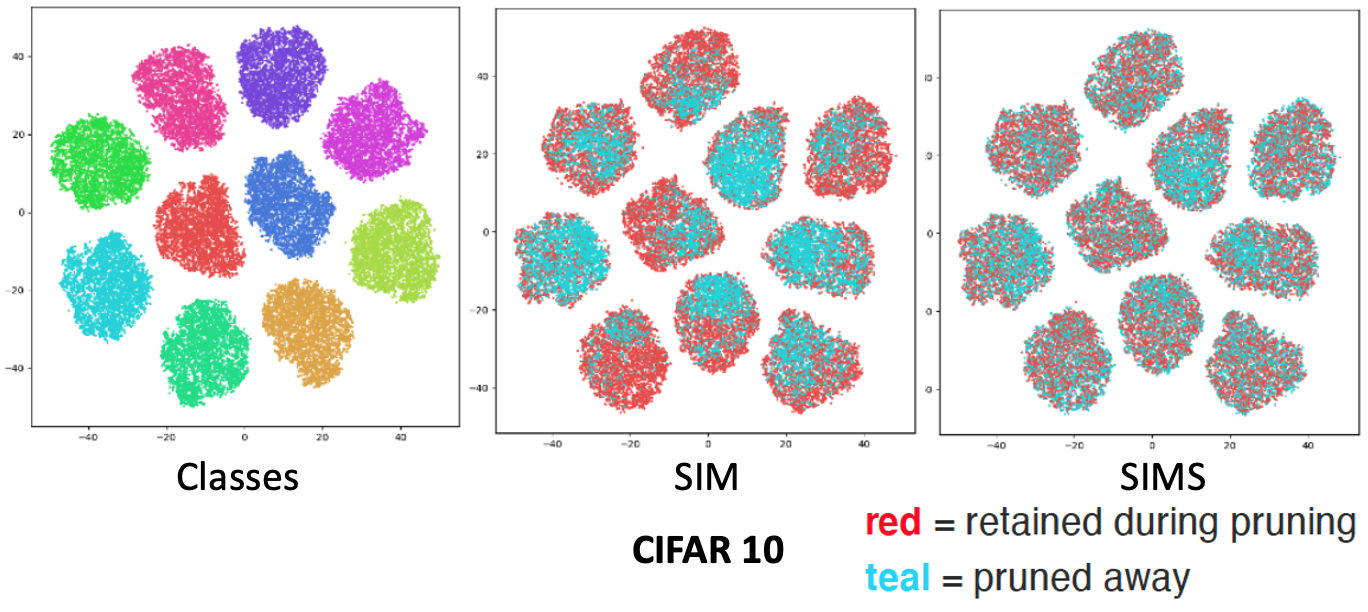}
\caption{T-SNE visualization for CIFAR-10 comparing SIM to SIMS.}
\vspace{-3em}
\label{fig:tsne_cifar10}
\end{center}
\end{figure}

The second ablation study was performed on the various components of the importance sampling procedure employed by SIMS. The results are given in Table~\ref{tab:sims_ablation} lower subsection. First, we observe a drastic decrease in performance without sampling (row 1) compared to any one of remaining four sampling procedures (row 2-5). 
Next, we analyze the effect of different variants of sampling by changing the within-class ratio, where 0\% means performing sampling on entire dataset, and 100\% means each class is equally pruned. From rows 2-5, we can see that as the difficulty and number of classes in the dataset increases, the benefit of sampling a percentage of the data within classes becomes more significant. In particular, we see about 2.1\% decrease in performance on Tiny-ImageNet when sampling strictly from the overall distribution compared to first sampling within each class and then sampling the remaining amount from the overall distribution (row 2 vs. row 3-5). The best overall performance across the three datasets is obtained with an initial 5\% sampling within class and remaining 95\% sampling from the overall distribution. To gain further insight into the benefit of importance sampling, we plotted pruned vs. retained samples on CIFAR-10 in the t-SNE space in Fig.~\ref{fig:tsne_cifar10}. Notice that importance sampling results in better coverage of each class distribution.
Additional results are in supplementary Fig.~ \ref{fig:tsne}.

\subsection{Cross-Model Generalization}
Finally, we conducted generalization experiments to test whether SIMS performs well when training downstream classification models on pruned datasets obtained via a different model architecture. From Table~\ref{tab:generalization}, SIMS performs better than Forgetting Scores with a ResNet50 model trained on the pruned datasets derived with a ResNet18 model. This suggests that some of the upfront costs of computing SIMS scores can be reduced by training expert models with a smaller architecture and subsequently training larger, downstream models on the pruned datasets.


\section{Conclusion}
We proposed a data pruning approach (\textit{a.k.a.} SIMS) which improves upon state-of-the-art pruning methods in terms of scalability and generalizability, as demonstrated on various datasets including the large-scale and long-tailed distribution dataset, iNaturalist. We showed that our method performs especially well at high pruning ratios and challenging fine-grained dataset compared to state-of-the-art. We showed that a key component of our proposed pruning procedure, importance sampling, was not only instrumental in improving the performance of the proposed SIM metric, but also can boost performance of other pruning metrics \textit{e.g.}, Forgetting Scores. Furthermore, we showed that SIMS can generalize better across unseen model architectures, which is desirable from practical perspective.
Our future direction is reducing the cost of training expert models required for achieving best results, integrating the full dataset performance as a prior in the importance sampling, and extending the metric to unsupervised learning of pruning models. Furthermore, even though we focused on the classification task in this paper, the idea of using SIM to characterize the data separability, data integrity, and model uncertainty is generic. The importance sampling process can also be extended to other tasks, such as detection and segmentation, since it is based on model uncertainty. Therefore, we believe the proposed SIMS is applicable to other tasks.




\bibliography{main}
\bibliographystyle{splncs04}





\end{document}


\title{Supplementary Materials: Data Pruning via Separability, Integrity, and Model Uncertainty-Aware Importance Sampling}

\author{Steven Grosz\inst{1}\thanks{This author's contribution was performed as an intern with Amazon.} \and
Rui Zhao\inst{2} \and
Rajeev Ranjan\inst{2} \and
Hongcheng Wang\inst{2} \and
Manoj Aggarwal\inst{2} \and
Gerard Medioni\inst{2} \and
Anil Jain\inst{1}
}
\authorrunning{S. Grosz et al.}
%
\institute{Michigan State University, East Lansing MI 48824, USA \email{\{groszste,jain\}@msu.edu} \and
Amazon, Seattle, WA 98101, USA \email{\{zhaori,rvranjan,hongchw,manojagg,medioni\}@amazon.com}}

\maketitle  
\section*{A. Prototype Scores with Combinations of Inter-Class and Intra-Class Sampling}

\begin{table}
\centering
\small{
\caption{Performance of pruning using the Prototype Score and retaining a percentage of the pruned dataset for samples at the center of each class cluster, rather than from the boundary.}
\label{tab:proto}
\begin{tabular}{lccc}
\hline
\multirow{2}{*}{Method Name} & \multicolumn{3}{c}{CIFAR-10} \\
& avg. $\alpha$=0.1-0.5 & avg. $\alpha$=0.5-0.9 & avg. $\alpha$=0.1-0.9 \\
\hline
Prototype & 93.10 & 79.25 & 85.48 \\
Prototype w/ 1\% center retained & \textbf{93.11} & 78.96 & 85.35  \\
Prototype w/ 5\% center retained & 92.89 & 82.38 & 87.19 \\
Prototype w/ 10\% center retained & 92.80 & \textbf{83.19} & \textbf{87.61} \\
Prototype w/ 100\% center retained & 90.67 & 82.27 & 86.26 \\
\hline
\end{tabular}
}
\end{table}

\section*{B. Pseudocode for SIMS}
Pseudocode for pruning with SIMS is given in Algorithm \ref{algo:data_pruning}.
\begin{algorithm}[h]
    \caption{Data Pruning with SIMS}
    \label{algo:data_pruning}
    
    \textbf{Require:} Dataset $\mathcal{D}=\{(x_i, y_i) \}_{i=1,...,|\mathcal{D}|}$, pruning ratio $\alpha$\\
    \textbf{Require:} $K$ randomly initialized expert models $\{{\theta_j^0}\}_{j=1,...,K}$ \\
    \textbf{Require:} Number of epochs $T$, Learning-rate scheduler $\{\eta_1, \ldots, \eta_T\}$ \\
    \textbf{Require:} Cross-entropy loss $\mathcal{L}$, SGD optimizer \\
    \textbf{Require:} Percentage of intra-class sampling $r$ \\
    \textbf{Output:} Pruned dataset $\mathcal{D}^\prime=\{(x_i, y_i) \}_{i=1,...,|\mathcal{D}^\prime|}$ \\

    \begin{algorithmic}[1]
        \STATE // Step 1: Optimize the parameters of each expert model on $\mathcal{D} $
        \FOR{$j$ in $\{1,...,K\}$}
            \STATE $\theta_{j} \gets SGD(\mathcal{L}(\mathcal{D}, \theta_j^0, T, \{\eta_1,...,\eta_T\}))$
        \ENDFOR
        \newline

        \STATE // Step 2: Calculate Separability $s(\cdot)$, Integrity $e(\cdot)$, Model Certainty $c(\cdot)$ and $SIM(\cdot)$ scores 
        \FOR{$i$ in $\{1,...,|\mathcal{D}|\}$}
            \STATE $s(x_i) \gets  \frac{1}{K}\sum_{j=1}^{K} s_{\theta_j}(x_i)$
            \STATE $e(x_i) \gets  \frac{1}{K}\sum_{j=1}^{K} e_{\theta_j}(x_i)$
            \STATE $c(x_i) \gets  1 - JSD(\{h_j\}_{j=1}^K\mid x_i)$
            \STATE $g(x_i) \gets \sqrt{(1-s(x_i))^2+c(x_i)^2}-\sqrt{(1-s(x_i))^2+(1-c(x_i))^2}$
            \STATE $SIM(x_i) \gets \sqrt{g(x_i)^2+e(x_i)^2}$
        \ENDFOR
        \newline

        \STATE // Step 3: Calculate importance weight $w(\cdot)$ for sampling
        \STATE $p(x) \sim \mathcal{N}(\mu_0, \sigma_0^2| \mathcal{D})$
        \STATE $t \gets \frac{1}{2}\left(\sin(\alpha \pi - \frac{\pi}{2}) + 1\right)$
        \STATE $\mu \gets F^{-1}\left(t; \mu_0, \sigma_0\right)$
        \STATE $\sigma \gets \alpha\sigma_0$
        \STATE $q(x|\alpha) \sim \mathcal{N}(\mu, \sigma^2)$
        \STATE $w(x|\alpha) \gets \frac{q(x|\alpha)}{p(x)}$
        \newline
        
        \STATE // Step 4: Perform importance sampling with $r$ ratio of intra-class sampling
    \end{algorithmic}
    \textbf{Return} $\mathcal{D}^\prime \leftarrow \text{ImportanceSample}(SIM(\cdot)|w(\cdot),r)$
\end{algorithm}

\section*{C. Qualitative Results}
\label{Qualitative Results}
\begin{figure}
\begin{center}
\includegraphics[width=0.9\linewidth]{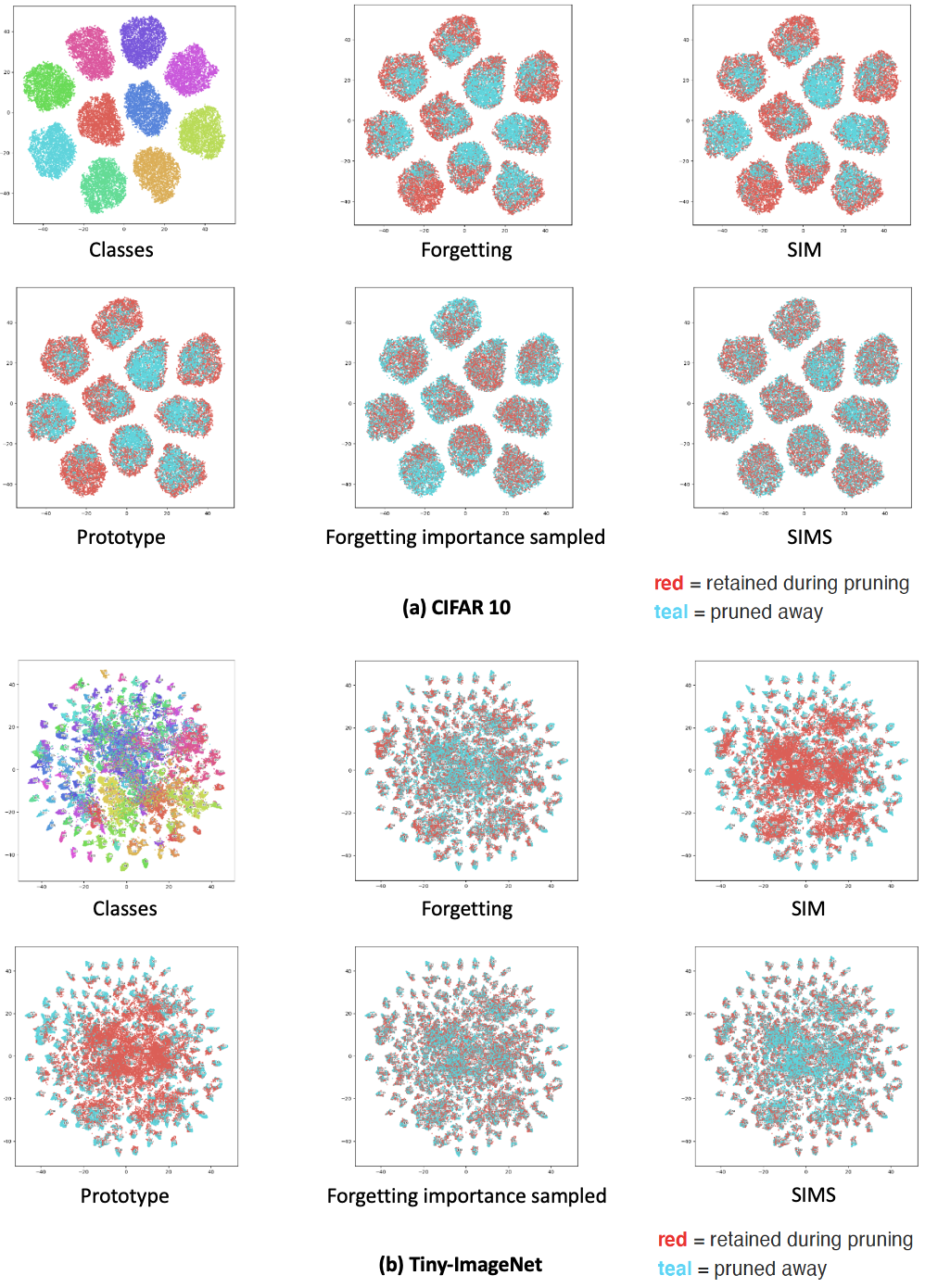}
\caption{t-SNE visualization.}
\label{fig:tsne}
\end{center}
\end{figure}

We conducted further qualitative analysis to gain deeper insights into the behavior of different pruning metrics. First, we generated visual representations by plotting embeddings of data samples that were either pruned or retained for a pruning ratio of 0.5 on CIFAR-10 and Tiny-ImageNet (Figure~\ref{fig:tsne}). The metrics considered in this analysis included Forgetting, SIM, Prototype, Forgetting with importance sampling, and SIMS. The samples depicted in blue represent those that were pruned, while the samples in red represent those that were retained.

\begin{figure*}
\begin{center}
\includegraphics[height=1.25\linewidth]{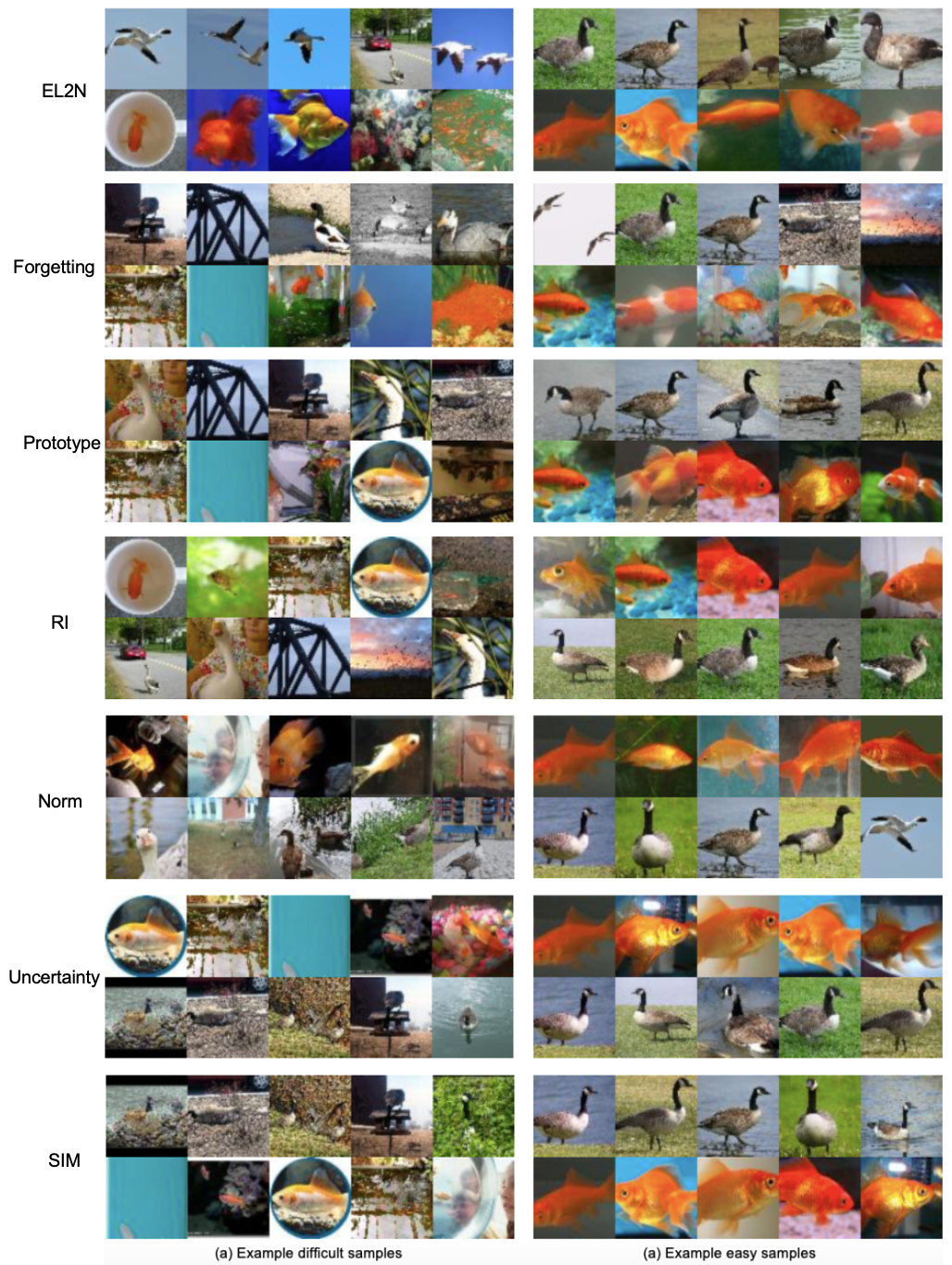}
\caption{Example high and low scoring images of goldfish and goose categories of the Tiny-ImageNet dataset according to various pruning metrics.}
\label{fig:ex_imgs}
\end{center}
\end{figure*}

For an easy dataset like CIFAR-10, the retained samples for the SIM and Prototype metrics tend to belong to the border of the class clusters, whereas for Tiny-ImageNet, the retained samples appear to be clustered in areas of high overlap between classes. Secondly, the effect of importance sampling across both datasets was a more uniform sampling of border vs. cluster center samples. To further understand the differences among the various pruning metrics, we have provided example images in Figure~\ref{fig:ex_imgs} of two classes (goldfish and goose) from the Tiny-ImageNet dataset which are high and low scoring for each of the pruning metrics. 

